\documentclass[final]{cvpr}

\usepackage{times}
\usepackage{epsfig}
\usepackage{graphicx}
\usepackage{amsmath}
\usepackage{amssymb}
\usepackage{soul}
\usepackage{color}
\usepackage{pifont}
\usepackage[ruled,vlined]{algorithm2e}
\usepackage{multirow}
\usepackage{booktabs}
\usepackage[skip=5pt]{caption} 

\usepackage[table,x11names]{xcolor}
\usepackage{enumitem}

\definecolor{mygray}{gray}{0.95}
\definecolor{myred}{rgb}{1.0, 0.0, 0.0}

\newcommand{\mbf}[1]{\mathbf{#1}}

\newcommand{\x}{\mathbf{x}}
\newcommand{\y}{\mathbf{y}}
\newcommand{\z}{\mathbf{z}}
\newcommand{\f}{\mathbf{f}}
\newcommand{\exit}{\mathbf{g}}
\newcommand{\policy}{\mathbf{\pi}}

\newcommand{\cmark}{\ding{51}}%
\newcommand{\xmark}{\ding{55}}%

\newcommand{\hlc}[2][yellow]{ {\sethlcolor{#1} \hl{#2}} }
\newcommand\bb[1]{\hlc[cyan]{Babak: #1}}

\newcommand{\partitle}[1]{\noindent\textbf{#1}}

\newcommand{\methodname}{FrameExit}
\def\etal{\emph{et al}\onedot}
\DeclareMathOperator*{\argmin}{arg\,min}

\usepackage[pagebackref=true,breaklinks=true,letterpaper=true,colorlinks,bookmarks=false]{hyperref}

\def\cvprPaperID{9966} 
\def\confYear{CVPR 2021}

\begin{document}

\title{FrameExit: Conditional Early Exiting for Efficient Video Recognition}

\author{Amir Ghodrati\thanks{Equal contribution} \qquad Babak Ehteshami Bejnordi\footnotemark[1] \qquad Amirhossein Habibian\\
{Qualcomm AI Research\thanks{ Qualcomm AI Research is an initiative of Qualcomm Technologies, Inc}}\\
{\tt\small \{ghodrati, behtesha, ahabibia\}@qti.qualcomm.com}


}

\maketitle

 \begin{abstract}
  In this paper, we propose a conditional early exiting framework for efficient video recognition. While existing works focus on selecting a subset of salient frames to reduce the computation costs, we propose to use a simple sampling strategy combined with conditional early exiting to enable efficient recognition. Our model automatically learns to process fewer frames for simpler videos and more frames for complex ones. To achieve this, we employ a cascade of gating modules to automatically determine the earliest point in processing where an inference is sufficiently reliable. We generate on-the-fly supervision signals to the gates to provide a dynamic trade-off between accuracy and computational cost. Our proposed model outperforms competing methods on three large-scale video benchmarks. In particular, on ActivityNet1.3 and mini-kinetics, we outperform the state-of-the-art efficient video recognition methods with $1.3\times$ and $2.1\times$ less GFLOPs, respectively. Additionally, our method sets a new state of the art for efficient video understanding on the HVU benchmark.

 \end{abstract}

\section{Introduction}

With the massive growth in the generation of video content comes an increasing demand for efficient and scalable action or event recognition in videos. The astounding performance of deep neural networks for action recognition~\cite{carreira2017quo, tran2018closer, zhou2018temporal, feichtenhofer2019slowfast, tran2019video, yue2015beyond} are obtained by densely applying 2D~\cite{yue2015beyond, lin2019tsm, zhou2018temporal, fernando2016rank} or 3D~\cite{tran2015learning, carreira2017quo, hara2017learning, feichtenhofer2019slowfast} models over video frames. Despite demonstrating top-notch performance in recognizing complex and corner case actions, the high data volumes, compute demands, and latency requirements, limit the application of the state-of-the-art video recognition models on resource-constrained devices.


\begin{figure}[t]
\centering
\includegraphics[width=0.98\linewidth]{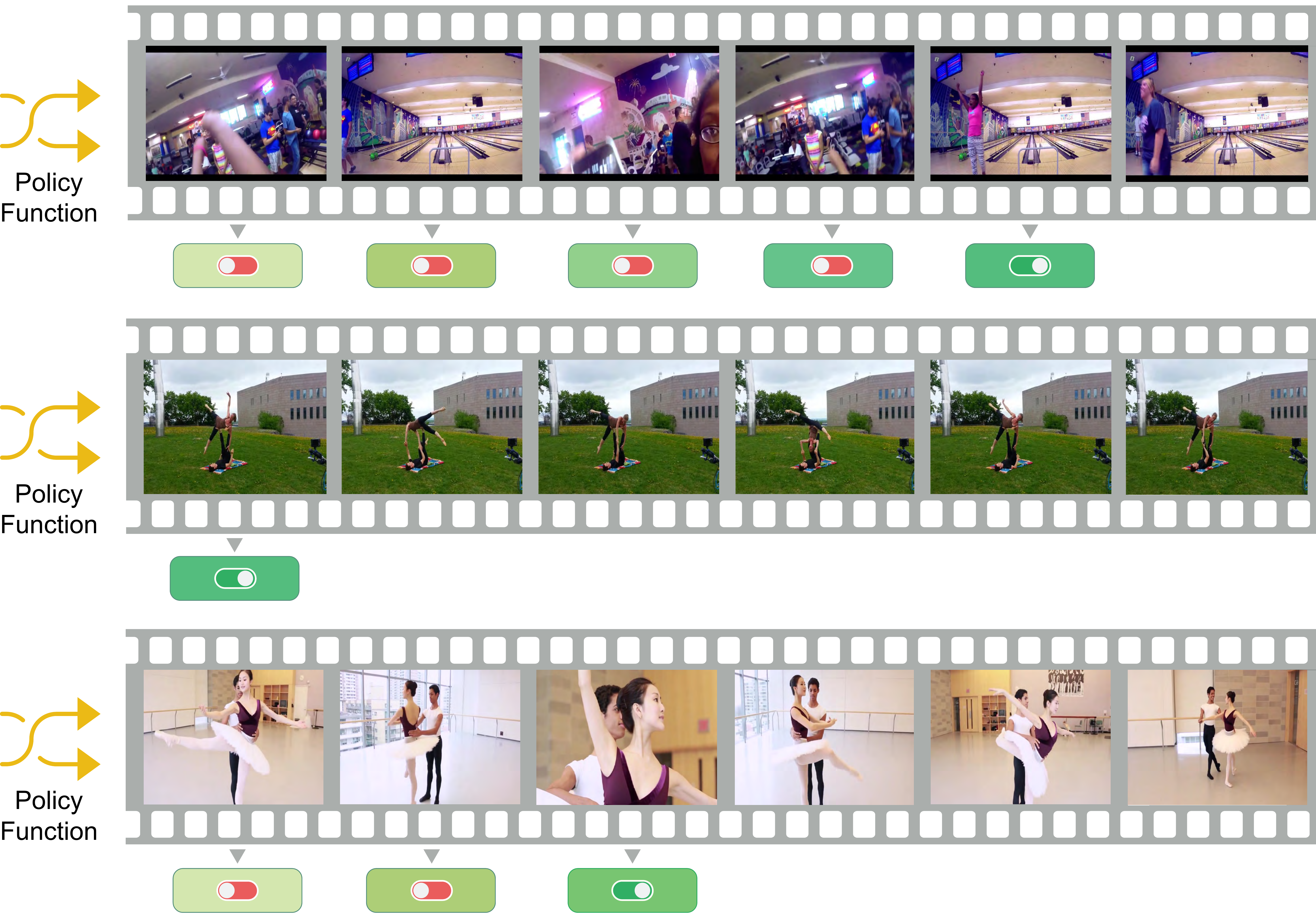}
\caption{{\bf Efficient video recognition by early exiting}. Our proposed method adjusts the amount of computation to the difficulty of the input, allowing for significant reduction of computational costs. Videos are adopted from~\cite{yt_video_qualitative1, yt_video_teaser1, yt_video_teaser3}.}
\label{fig:teaser}
\vspace*{-5mm}
\end{figure}

Extensive studies have been conducted to remedy this issue by designing efficient and light-weight architectures~\cite{piergiovanni2019tiny, feichtenhofer2020x3d, tran2018closer, qiu2017learning, xie2017rethinking, zolfaghari2018eco, tran2019video, kopuklu2019resource, lin2019tsm, fan2019more}. These models have a static computational graph and treat all the videos equally regardless of how complex or easy they are for recognition and hence yield sub-optimal results. A large body of research has been focusing on selecting a subset of salient frames to efficiently process the video conditioned on the input~\cite{yeung2016end, wu2019adaframe, zheng2020dynamic, fan2018watching, gao2020listen, korbar2019scsampler}.
Current methods for frame selection rely on learning a policy function to determine what action should be taken on the selected frame (e.g. process by a heavy recognition model~\cite{wu2019liteeval}, process at a specific spatial resolution~\cite{meng2020ar}, etc.). Most of these methods either rely on the assumption that salient frames for the sampler network are also salient for the recognition network~\cite{korbar2019scsampler, gao2020listen} or require carefully selected reward functions in case of using policy gradient methods~\cite{yeung2016end, wu2019adaframe, wu2019multi}. Moreover, the sampler network may create an additional computational overhead to the model.

An alternative promising direction to reduce the computational complexity of analyzing video content is conditional compute using early exiting. Early exiting has recently been explored for image classification tasks by inserting a cascade of intermediate classifiers throughout the network \cite{lee2015deeply, branchynet, yang2020resolution, multiscaledense}.
In this line of work, the model adjusts the amount of computation to the difficulty of the input and allows for significant reduction of computational requirements. Inspired by that, we design an efficient video recognition model that performs automatic early exiting by adjusting the computational budget on a per-video basis.
Our motivation is that a few frames are sufficient for classifying ``easy'' samples, and only some ``hard'' samples need temporally detailed information (see Figure~\ref{fig:teaser}).


In this paper, we propose \methodname, a conditional early exiting framework with learned gating units that decide to stop the computation when an inference is sufficiently reliable. \methodname~has $T$ classifiers accompanied by their associated gates that are attached at different time steps to allow early exiting. The gates are learned in a self-supervised fashion to control the trade-off between model accuracy and total computation costs. We use the recognition loss as a proxy to generate on-the-fly pseudo-labels to train the gates.

Additionally, our early exiting mechanism combined with a simple, deterministic sampling strategy obviates the need for complex sampling policy functions and yet achieves excellent recognition performance. Finally, we propose an accumulated feature pooling module to generate video representations that enable more reliable predictions by the model. Our contributions are as follows:

\begin{itemize}
\item We propose a method that employs a simple, deterministic frame sampling strategy, combined with an accumulated feature pooling module to obtain accurate action recognition results. 
\item We propose a conditional early exiting framework for efficient video recognition. We use a cascade of gating modules to determine when to stop further processing of the video. The gates adapt the amount of computation to the difficulty of an input video, leading to significant reductions in computational costs.

\item We show state-of-the-art performance on three large-scale datasets. In all cases, we greatly improve the inference efficiency at a better or comparable recognition accuracy. In particular, on the HVU~\cite{diba2020large} dataset, we report $5\times$ reduction in computation costs while improving the recognition accuracy upon the state-of-the-art methods.
\end{itemize}

\section{Related work}
\partitle{Efficient Video Recognition:}
There are two lines of approaches for efficient video recognition. The first focuses on proposing new lightweight video architectures. This can be obtained by decomposing 3D filters into separate 2D spatial and 1D temporal filters~\cite{tran2018closer, qiu2017learning, xie2017rethinking}, extending efficient 2D architectures to their 3D counterparts~\cite{tran2019video, kopuklu2019resource}, using shifting operations~\cite{lin2019tsm, fan2020rubiksnet}, or exploring neural architecture search for videos~\cite{piergiovanni2019tiny, feichtenhofer2020x3d}. 
Our method is agnostic to the network architecture and can be complementary to these types of methods.

The second line of works focus on saving compute by selecting salient frames/clips using a policy function parametrized with a neural network (CNN or RNN). This is commonly done by training agents to find which frame to observe next~\cite{yeung2016end, wu2019adaframe, zheng2020dynamic, fan2018watching}, ranking frames based on saliency~\cite{korbar2019scsampler}, gating frames~\cite{hussein2020timegate}, skipping RNN states~\cite{campos2017skip}, or by using audio to select relevant frames~\cite{gao2020listen, korbar2019scsampler}. 
LiteEval~\cite{wu2019liteeval} proposes binary gates for selecting coarse or fine features. At each time step, LiteEval computes coarse features with a lightweight CNN to determine whether to examine the current frame more carefully using a heavy CNN. Meng~\etal.~\cite{meng2020ar} propose to adaptively select the input resolution, on a per-frame basis to further balance the accuracy vs compute. However, using policy networks in these methods may come with additional memory and compute overhead. Moreover, optimization is sub-optimal if policy and recognition networks are not trained jointly~\cite{korbar2019scsampler,gao2020listen}, or requires carefully selected reward functions if policy gradient methods are used~\cite{wu2019adaframe, yeung2016end, wu2019multi}.
In contrast, our method relies on a single recognition network and does not require complex RL optimization for frame sampling. We formulate the problem in the early-exiting framework and show that a simple sampling strategy, if combined with a proper exiting function, leads to excellent recognition performance.

\begin{figure*}[htb!]
\centering
\includegraphics[width=0.98\linewidth]{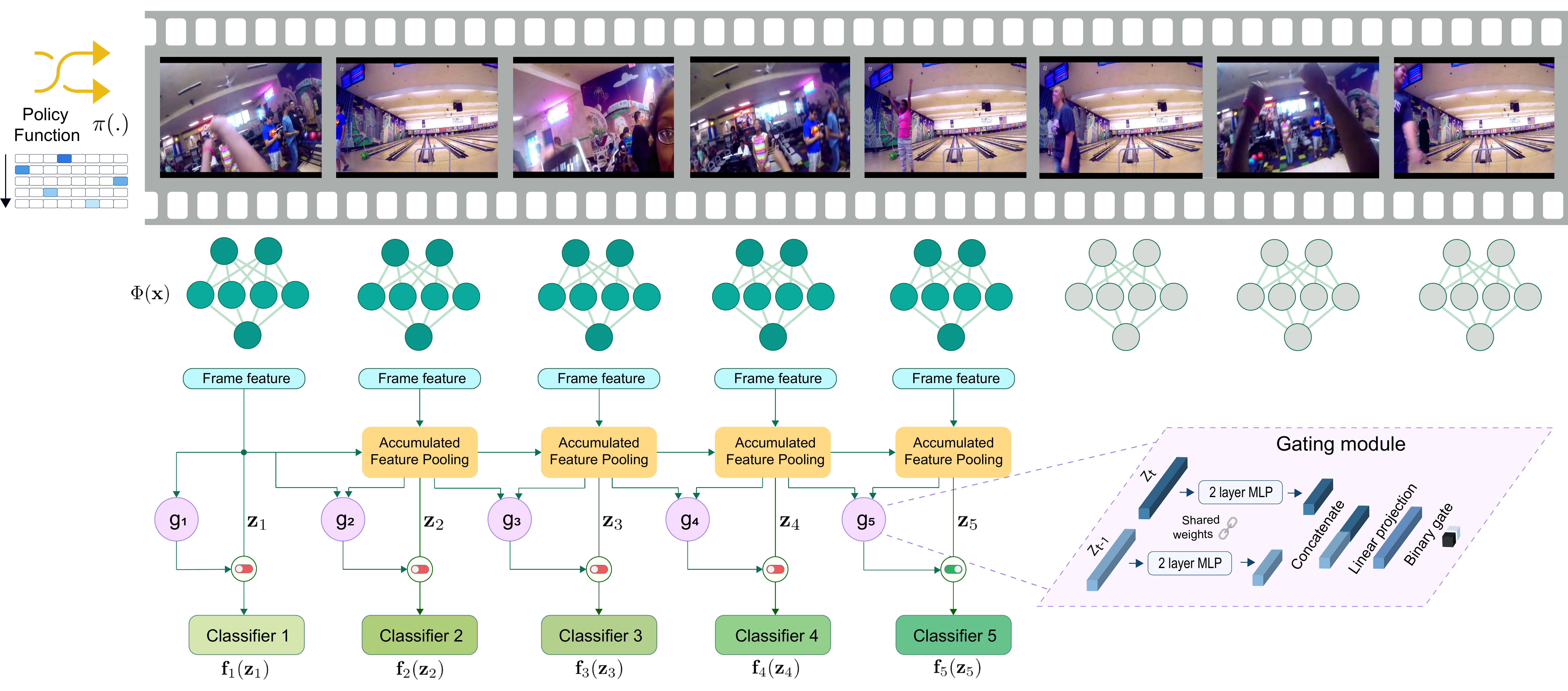}
\caption{{\bf The overview of \methodname.} Given a video, at each time step $t$, we sample a frame from the video using the deterministic policy function $\policy$. Each frame is independently represented by the feature extraction network $\Phi$ and is aggregated to features of previous time steps using the accumulated feature pooling module (for $t>1$). The gating modules ($g_t$) are trained to allow the network to automatically determine the earliest exiting point based on the inferred complexity of the input video. The architecture of the gating module is illustrated in the bottom right corner of the figure. Note that $g_1$ only receives $z_1$ as input. The video is adopted from~\cite{yt_video_teaser1}.}
\label{fig:overview}
\vspace*{-3mm}
\end{figure*}

\medskip\partitle{Conditional compute via early exiting:} Conditional computation in neural networks aims to dynamically allocate the components of the model (e.g. layers, sub-networks, etc.) on a per-input basis. ConvNet-AIG \cite{convnetaig}, SkipNet \cite{skipnet}, and BlockDrop \cite{wu2018blockdrop} exploit the robustness of Residual Networks (ResNets) to layer dropping and activate or deactivate full residual blocks conditioned on the input, to save compute. GaterNet \cite{gaternet}, dynamic  channel pruned networks \cite{dynamicchannelpruning}, batch-shaped channel-gated networks \cite{bejnordi2019batch}, and Conditional channel gated networks for continual learning \cite{abati2020conditional} perform gating in a more fine-grained fashion by turning individual channels of a layer on or off conditioned on the input. Other works focus on adaptive spatial attention for faster inference \cite{figurnov2017spatially, verelst2020dynamic, wang2020learning}. Concurrent to our work, AdaFuse \cite{meng2021adafuse} dynamically fuses channels from current and history feature maps to reduce the computation. 
A major research direction in conditional compute is early exiting. Prior works have mainly used early exiting for image classification. To adapt computational costs to the difficulty of the input, Deeply-Supervised Nets \cite{lee2015deeply} and 
BranchyNet \cite{branchynet} propose architectures that are composed of a cascade of intermediate classifiers. This allows simpler examples to exit early via an intermediate classifier while more difficult samples proceed deeper in the network for more accurate predictions. Multi-scale dense networks \cite{multiscaledense} and adaptive resolution networks \cite{yang2020resolution} focus on spatial redundancy of input samples and use a multi-scale dense connection architecture for early exiting.

Our model draws inspiration from these works, but rather than having intermediate classifiers operating on intermediate network features, we focus on early exiting over the temporal dimension for video recognition.


\section{\methodname}
\label{sec:method}

Given a set of videos and their labels $\{\mbf v_i, \y_i\}_{i=1}^D$, we aim to classify each video by processing the minimum number of frames. Figure~\ref{fig:overview} illustrates the overall architecture of our proposed model. Our model consists of \textit{i)} a frame sampling policy $\policy$, \textit{ii)} a feature extraction network $\Phi$, \textit{iii)} an accumulated feature pooling module, and \textit{iv)} $T$ classifiers $\f_t$ and their associated exiting gates $\exit_t$, where $T$ is the number of input frames.
Given an input video, we extract a partial clip $\x_{1:t}$ by sampling $t$ frames, one at a time, from the video based on the sampling policy $\policy$:
%
\begin{equation}
    \x_{1:t} = [\x_{1:t-1}; \x_t], \quad \x_t = \mbf v_{\policy(t)},
\end{equation}

where $\x_{1:t-1}$ denotes a partial clip of length $t-1$ and $\x_t$ is a single video frame. 
We use the feature extraction network $\Phi$ to generate independent representation for each frame $\x_t$.
These representations are then aggregated using the accumulated feature pooling module. The resulting clip level representation, $\z_t$, is then passed to the classifier $\f_t$ and its associated early exiting gate $\exit_t$.

Starting from a single-frame clip, we incrementally add more temporal details at each time step until one of the gates predicts the halt signal. Each gate $\exit_t: (\z_{t-1},\z_{t}) \rightarrow \{0,1\}$ is a binary function indicating whether the network has reached a desired confidence level to exit. A gate receives the aggregated features $\z_{t}$ and $\z_{t-1}$ as input. This allows the gate to make a more informed decisions by considering the agreement between temporal features. For example, a highly confident incorrect exiting prediction solely based on the current frame representation $\z_t$ could potentially be mitigated if there is a significant disagreement with the representation $\z_{t-1}$. Note that the first gating module in the network only receives $\z_{1}$ as input. In the end, the final video label will be predicted as:
%
%

%
\vspace{-3mm}
\begin{equation}
    \y = \f_t(\z_{t}), \quad\text{where}~t=\argmin_t\{t~|~\exit_t=1\}
\end{equation}
\vspace{-3mm}

where $t$ is the earliest frame that meets the gating condition. Note that if none of the gates predict the halt signal, the last classifier will classify the example. In the following sections, we describe our choice for frame sampling policy, accumulated feature pooling module, and our gates for early exiting.

\medskip\partitle{Frame Sampling Policy.} 
\label{sec:method:sampling}
We sample $T$ frames, one at a time, using the policy function $\pi(t)$. While in most existing works $\policy$ is parameterized with a light-weight network and is trained using policy gradient methods~\cite{sutton2018reinforcement} or Gumbel reparametrization~\cite{maddison2016concrete}, we use a simple, deterministic, and parameter-free function and show that it performs as well as sophisticated frame selection models. Our sampling function $\policy$ follows a coarse-to-fine principle for sampling in temporal dimension. It starts sampling from a coarse temporal scale and gradually samples more frames to add finer details to the temporal structure. Specifically, we sample the first frames from the middle, beginning, and end of the video, respectively, and then repeatedly sample frames from the halves (check Appendix for more details). Compared to sampling sequentially in time, this strategy allows the model to have access to a broader time horizon at each timestamp while mimicking the behaviour of RL-based approaches that jumps forward and backward to seek future informative frames and re-examine past information. 


\medskip\partitle{Feature extraction network.} 
\label{sec:method:backbone}
Our feature extraction network $\Phi(\x_i;\theta_\Phi)$ is a 2D image representation model, parametrized by $\theta_\Phi$, that extracts features for input frame $\x_i$. We use ResNet-50 \cite{he2016deep}, EfficientNet-b3 \cite{tan2019efficientnet}, and X3D-S \cite{feichtenhofer2020x3d} in our experiments.


\medskip\partitle{Accumulated Feature Pooling.} This step aims to create a representation for the entire clip $\x_{1:t}$. A major design criteria for our accumulated feature pooling module is efficiency. To limit the computation costs to only the newly sampled frame, the clip representation is incrementally updated.
Specifically, given the sampled frame $\x_t$ and features $\z_{t-1}$, we represent a video clip as:
\begin{equation}
    \z_{t}= \Psi(\z_{t-1}, \Phi(\x_i;\theta_\Phi)),
    \label{eq:aggregation}
\end{equation}
 where $\Psi$ is a temporal aggregation function that can be implemented by statistical pooling methods such as average/max pooling, LSTM~\cite{hochreiter1997long, ghodrati2018video}, or self-attention~\cite{vaswani2017attention}.

\medskip\partitle{Early Exiting.} While processing the entire frames of a video is computationally expensive, processing a single frame may also restrict the network's ability to recognize an action. Our conditional early exiting model has $T$ classifiers accompanied by their associated early exiting gates that are attached at different time steps to allow early exiting. Each classifier $\f_t$ receives the clip representation $\z_t$ as input and makes a prediction about the label of the video. During training, we optimize the parameters of the feature extractor network and the classifiers using the following loss function:
\begin{equation}
\mathcal{L}_{cls} = 
\frac{1}{T} \sum_{t=0}^{t=T} \ell_{cls}(\f_t(\z_t;\theta_f), \y)
\end{equation}

In our experiments, we use the standard cross-entropy loss for single-label video datasets and binary cross-entropy loss for the multi-label video datasets.

We parameterize each exiting gate $\exit_t$ as a multi-layer perceptron, predicting whether the partially observed clip $\x_{1:t}$ is sufficient to accurately classify the entire video. The gates have a very light design to avoid any significant computational overhead. As mentioned earlier, each gate $\exit_t(\z_{t-1},\z_{t}) \rightarrow \{0,1\}$ receives as input the aggregated representations $\z_{t}$ and $\z_{t-1}$ (See Figure~\ref{fig:overview} - bottom right). 
Each of these representations are first passed to two layers of MLP with 64 neurons independently (shared weights between the two streams). The resulting features are then concatenated and linearly projected and fed to a sigmoid function. Note that $g_1$ only receives $\z_1$ as input.

During training, gates may naturally learn to postpone exiting so that the last classifier always generates the model output because that may tend to maximize accuracy at the expense of additional processing. However, this sort of training largely defeats the purpose of the early exiting architecture. To overcome this problem we regularize the gates such that they are enforced to early exit. The parameters of the gates $\theta_g$ are learned in a self-supervised way by minimizing the binary cross-entropy between the predicted gating output and pseudo labels $\y^{g}_{t}$: 
\begin{equation}
    \mathcal{L}_{gate}= \frac{1}{T} \sum_{t=0}^{t=T}\operatorname{BCE}(\exit_t(\z_{t-1},\z_{t};\theta_g), \y^{\exit}_{t}),
\end{equation}

We define the pseudo labels for the gate $\exit_t$ based on the classification loss:
\vspace{-1mm}
\begin{equation}
  \y^{g}_{t} =
    \begin{cases}
      1 & \text{$\ell_{cls}(\f_t(\z_{t}), \y)\leq \epsilon_t$} \\
      0 & \text{else}\\
    \end{cases}       
\end{equation}
\vspace{-1mm}


where $\epsilon_t$ determines the minimum loss required to exit through $\f_t$. A label $1$ indicates the exiting signal while label $0$ indicates that the network should proceed with processing further frames. Using this loss, we enforce the gate $\exit_t$ to exit when the classifier $\f_t$ generates a sufficiently reliable prediction with a low loss. Provided that the early stage classifiers observe very limited number of frames, we only want to enable exiting when the classifier is highly confident about the prediction, i.e. $\f_t$ has a very low loss. Hence, it is preferred to use smaller $\epsilon_t$ for these classifiers. On the other hand, late stage classifiers mostly deal with difficult videos with high loss. Therefore, as we proceed to later stage classifiers, we increase $\epsilon_t$ to enable early exiting. In our experiments, we define $\epsilon_t=\beta \exp(\frac{t}{2})$, where $\beta$ is a hyper-parameter that controls the trade-off between model accuracy and total computation costs. The higher the $\beta$, the more computational saving we obtain.

The final objective for training our model is given as:
\begin{equation}
    \mathcal{L} = \mathbb{E}_{(\mbf v,\y)\sim D_{train}}\left[\mathcal{L}_{cls} + \mathcal{L}_{gate}\right]
\end{equation}

Note that we use equal weights for the classification and gating loss terms in all of our experiments. 

\section{Experiments}
\label{sec:exp}

\begin{figure}[t]
\centering
\includegraphics[width=1\linewidth]{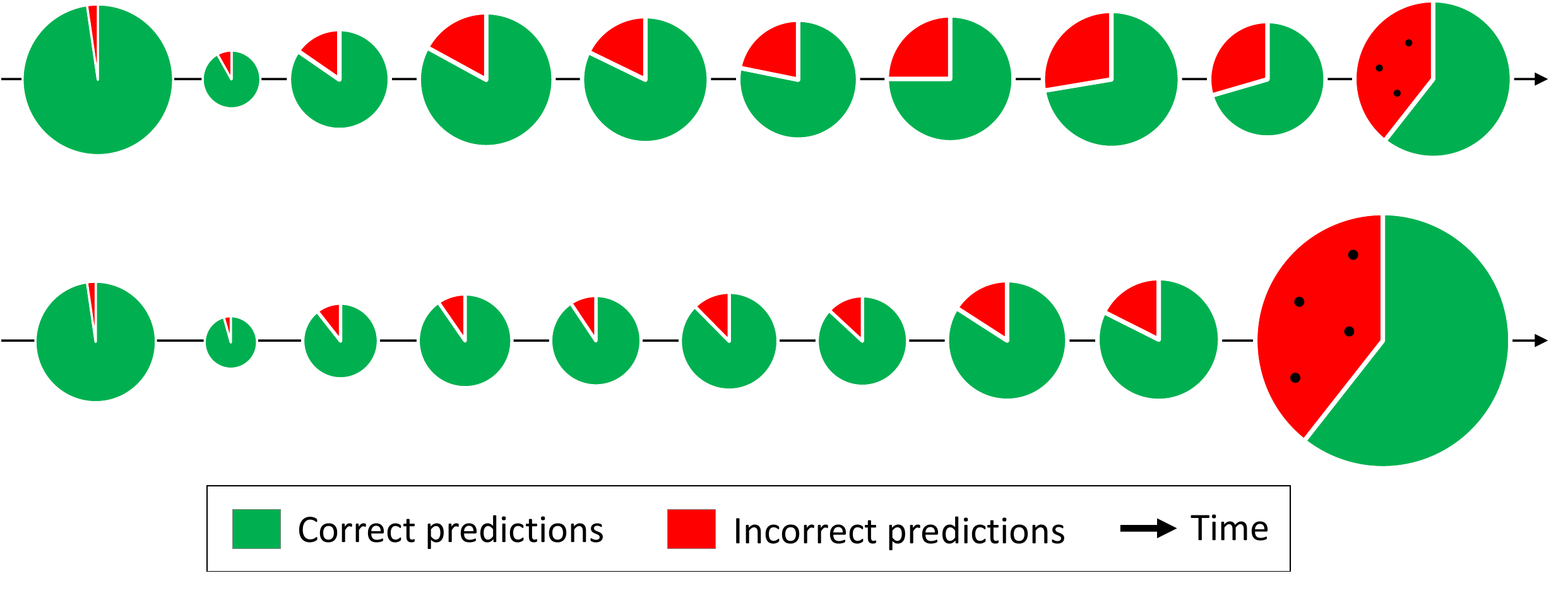}
\caption{\textbf{Watermelon visualization of \methodname~ predictions over time.} The exiting statistics of our method over time for $\beta=1e^{-4}$ (top) and $\beta=1e^{-6}$ (bottom) on the validation set of ActivityNet. The area of a circle at time step $t$ represents the percentage of samples that exited through classifier $t$. Easier examples exit earlier from the network with higher accuracy while only hard examples reach to late stage classifiers leading to increased miss-classifications. The illustration is inspired from~\cite{wu2019adaframe}.}
\label{fig:percentage}
\vspace*{-5mm}
\end{figure}

We conduct extensive experiments to investigate the efficacy of our proposed method using three large-scale datasets on two video understanding tasks, namely action recognition and holistic video understanding, as described in Section~\ref{sec:exp:setup}. Our experiments demonstrate that \methodname~outperforms the state of the art while significantly reducing computational costs as discussed in Section~\ref{sec:exp:main}. Finally, we present an ablation study of our design choices in Section~\ref{sec:exp:ablation}

\subsection{Experimental setup}
\label{sec:exp:setup}
\textbf{Datasets.}
\label{sec:exp:datasets}
We conduct experiments on three large-scale datasets: \textit{ActivityNet-v1.3}~\cite{caba2015activitynet}, \textit{Mini-Kinetics}~\cite{kay2017kinetics},  and \textit{HVU}~\cite{diba2020large}. ActivityNet-v1.3~\cite{caba2015activitynet} is a long-range action recognition dataset consisting of $200$ classes with $10,024$ videos for training and $4,926$ videos for validation. The average length of the videos in this dataset is $167$ seconds. 
Mini-Kinetics~\cite{kay2017kinetics} is a short-range action recognition dataset, provided by~\cite{meng2020ar}, that contains $200$ classes from Kinetics dataset~\cite{kay2017kinetics} with $121,215$ training and $9,867$ testing videos. The average duration of videos is $10$ seconds.
HVU~\cite{diba2020large} is a large-scale, multi-label dataset for holistic video understanding with $3142$ classes of actions, objects, scenes, attributes, events and concepts. It contains $476k$ training and $31k$ validation videos. The average duration of videos in this dataset is $8.5$ seconds.

\medskip\textbf{Evaluation metrics.} 
\label{sec:exp:metrics}
Following the literature, we use mean average precision (mAP) and top-1 accuracy to evaluate accuracy for multi-class (Mini-Kinetics) and multi-label (ActivityNet and HVU) classification respectively.
We measure the computational cost as giga floating point operations (GFLOPs). As a reference, ResNet-50 and EfficientNet-b3 have $4.12$ and $1.8$ GFLOPs respectively for a single input image of size $224 \times 224$. Moreover, the X3D-S architecture has $1.96$ GFLOPs for an input size of $13\times160\times160$. As different baseline methods use different number of frames for recognition, we report per video GFLOPs in all of our experiments.

\begin{figure}[t]
\centering
\resizebox{0.8\columnwidth}{!}{\includegraphics[width=1\linewidth]{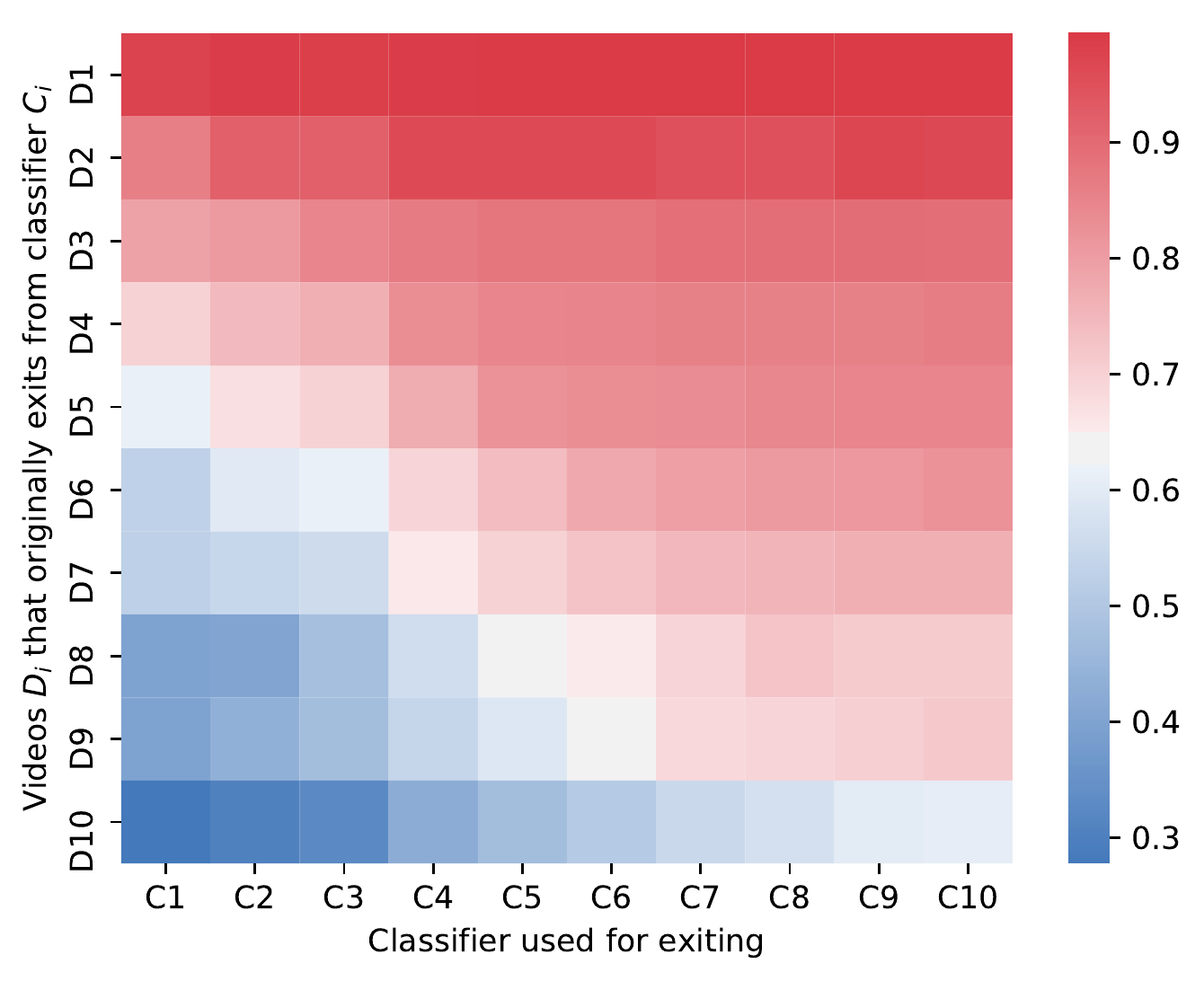}}
\caption{\textbf{Illustration of the exiting performance.} The matrix entry $(i,j)$ illustrates the accuracy of the classifier $C_j$, if it were to classify the video data $D_i$ that originally early exited by the model through $C_i$. Interestingly, if a video prematurely exits earlier than what was decided, we significantly lose accuracy while if it exits later than the model's decision, no significant gain is obtained. 
}
\label{fig:confusion}
\vspace*{-4mm}
\end{figure}

\medskip\textbf{Implementation details.} 
\label{sec:exp:implemetation_detail}
We use ResNet-50~\cite{he2016deep} and EfficientNet-b3~\cite{tan2019efficientnet} hereafter referred to as ResNet and EfficientNet, as well as X3D-S \cite{feichtenhofer2020x3d} as our backbone networks. Both backbones are pretrained on ImageNet. We remove the last classification layers from the backbone networks, and replace it with a fully-connected layer with $4096$ neurons. We set the number of input frames, $T$, to $10$. 
We use max-pooling as the accumulated feature pooling function, $\Psi$ in Eq.~\ref{eq:aggregation}, for all the experiments unless otherwise stated. We train the model in two stages: we first train the backbone network and classifiers, then we learn the parameters of the gating modules. We use Adam optimizer~\cite{adam} with a learning rate of $1e^{-4}$ for both training stages. The first training step runs for $35$ epochs whilst dropping the learning rate after $16$, and $30$ epochs with a factor of $0.1$. The second training step runs for $10$ epochs and the learning rate is dropped after epochs $5$ and $8$. We set the hyper-parameter $\beta$ ranging from $1e^{-6}$ to $1e^{-2}$ to generate varying trade-off points between accuracy and computational costs.

\subsection{Results}
\label{sec:exp:main}
We first analyze the behaviour of our conditional early exiting method. Then we compare \methodname~ with the state of the art in action recognition and holistic video understanding. This section is concluded by reporting several qualitative results.

\medskip\partitle{Conditional early exiting.} \label{sec:exp:watermelon} We analyze the effectiveness of \methodname~in adjusting the amount of computations per video based on its content. Figure~\ref{fig:percentage} illustrates predictions and the number of frames processed by two \methodname~ models trained with different values of $\beta=1e^{-4}$ and $\beta=1e^{-6}$. As expected, a higher $\beta$ encourages more video samples to exit early from the network (top row) compared to a lower $\beta$ (bottom row). A general observation is that the more we proceed to classifiers in later stages, the more inaccurate the predictions become. This may sound counter-intuitive, because if we were to have a model without early exiting, late-stage classifiers produce the most accurate predictions. 
However, the trend shown in Figure~\ref{fig:percentage} is highly desirable, because it shows that the easier examples have already exited from the network while only hard examples reach to late stage classifiers. 

This observation is more clear in Figure~\ref{fig:confusion}. The matrix entry $(i,j)$ illustrates the accuracy of the classifier $C_j$, if it were to classify the video data $D_i$ that originally early exited by the model through $C_i$. The diagonal entries $(i,j=i)$ in the matrix represent the actual early-exiting results obtained by the model. In an ideal early exiting scenario, for each row $i$, it is desired to have a significantly lower performance for $C_j$ compared to $C_i$ when $j<i$ and a similar accuracy to $C_i$ when $j > i$. A similar pattern is observed in Figure~\ref{fig:confusion}. As can be seen, if a video prematurely exits earlier than what was decided by the model, we lose accuracy significantly while if it exits later than the model's decision, no significant gain in accuracy is obtained. \methodname~learns to exit from the optimal classifier to balance computational costs and model accuracy.

\begin{figure}
\centering
\includegraphics[width=0.85\linewidth]{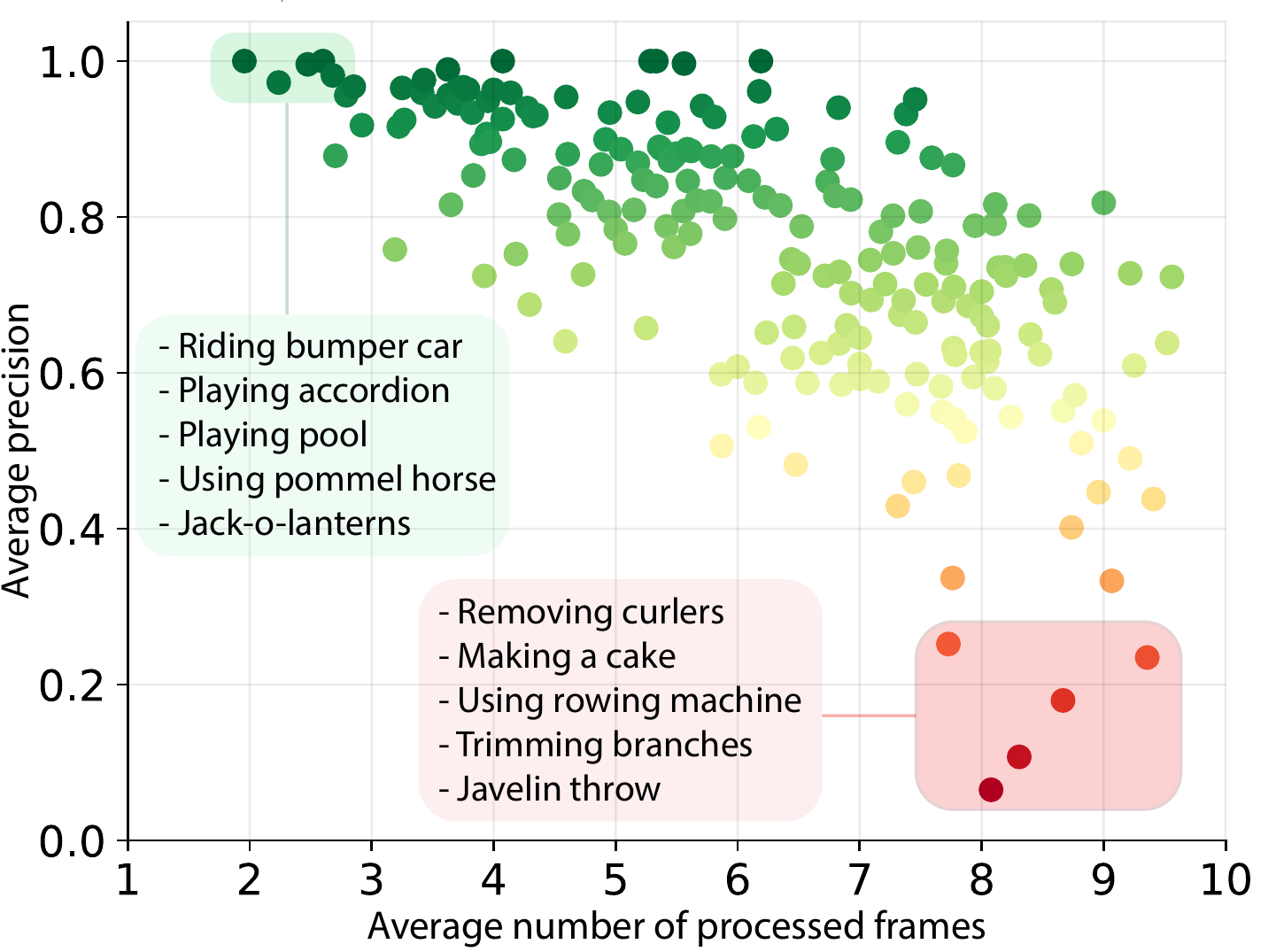}
\caption{\textbf{AP vs number of processed frames per category in the ActivityNet dataset.} Categories with object/scene cues exit in the early stages with high accuracy while complex actions require more frames for recognition.}
\label{fig:class_stat}
\end{figure}

Figure~\ref{fig:class_stat} illustrates average precision of all categories in the ActivityNet dataset as well as the average number of frames required for each category to confidently exit from the network. As can be seen, the classes with strong visual cues (such as riding bumper car, playing accordion, playing pool, etc.) appear to be simpler and therefore exit in the early stages with a high accuracy. In contrast, actions which involve more complex interactions and require a series of frames to unveil (such as Javelin throw and trimming branches) take more frames to be recognized. 

\medskip\partitle{Comparison to state of the art: Action recognition.}
\label{sec:exp:sota}
%
\begin{table}[t]
\caption{\textbf{Comparison to state of the art on action recognition.} \methodname~outperforms competing methods in terms of accuracy and efficiency using ResNet, EfficientNet, and X3D-S backbones. Results of other methods are adopted from~\cite{meng2020ar}. $^*$(IA$|$IA) denotes the variant of the ListenToLook method that additionally uses audio for sampling and recognition.}
\label{table:sota}
\centering
\resizebox{1.0\columnwidth}{!}
{
\begin{tabular}{|l|cc|cc|}
\hline
 \multicolumn{1}{|c|}{\multirow{2}{*}{}} &
 \multicolumn{2}{c|}{\cellcolor{mygray}\textbf{ActivityNet}} & \multicolumn{2}{c|}{\cellcolor{mygray}\textbf{Mini-kinetics}}\\ 
 \cline{2-5}
 
 & mAP (\%) & GFLOPs & Top-1 (\%) & GFLOPs \\
\hline
\rowcolor{mygray}\textit{ResNet} & & & &\\
AdaFrame~\cite{wu2019adaframe}              & 71.5 & 79.0 & -    & -  \\ 
LiteEval~\cite{wu2019liteeval}              & 72.7 & 95.1 & 61.0 & 99.0 \\
ListenToLook~\cite{gao2020listen}           & 72.3 & 81.4 & -    & -\\ 
ListenToLook (IA$|$IA)$^{\mbox{*}}$~\cite{gao2020listen}           & 75.6 & 37.5 & -    & -\\
SCSampler~\cite{korbar2019scsampler}        & 72.9 & 41.9 & 70.8 & 41.9\\ 
AR-Net~\cite{meng2020ar}                    & 73.8 & 33.5 & 71.7 & 32.0\\ 
\textbf{Ours (w/o exit)}                    & \textbf{77.3} & 41.2 & \textbf{73.3} & 41.2 \\
\textbf{Ours (\methodname)}                 & 76.1 & \textbf{26.1} & 72.8 & \textbf{19.7}\\
\hline
\rowcolor{mygray}\textit{EfficientNet} & & & &\\
AR-Net~\cite{meng2020ar}                    & 79.7 & 15.3 & 74.8 & 16.3\\
\textbf{Ours (w/o exit)}                    & \textbf{81.1} & 18.0 & \textbf{75.9} & 18.0 \\
\textbf{Ours (\methodname)}                 & 80.0 & \textbf{11.4} & 75.3 & \textbf{7.8}\\

\rowcolor{mygray}\textit{X3D-S} & & & &\\
\textbf{Ours (w/o exit)}                    & \textbf{87.4} & 19.6 & - & - \\
\textbf{Ours (\methodname)}                 & 86.0 & \textbf{9.8} & - & -\\

\hline
\end{tabular}
}
\end{table}
%

\noindent We compare our method with an extensive list of recent works on efficient video recognition: AR-Net~\cite{meng2020ar}, AdaFrame~\cite{wu2019adaframe}, LiteEval~\cite{wu2019liteeval}, SCSampler~\cite{korbar2019scsampler}, MARL~\cite{wu2019multi}, and ListenToLook~\cite{gao2020listen}.
AR-Net uses MobileNet-V2 as the sampler network and adaptively chooses a ResNet architecture with varying depths as the recognition network. The method is additionally evaluated with variants of EfficientNet~\cite{tan2019efficientnet} as the recognition network.
AdaFrame and LiteEval both use MobileNet-V2 as the policy network and ResNet-101 as the recognition network.
SCSampler uses MobileNet-V2 as the sampler network and ResNet-50 as the recognition network. 
MARL uses a ResNet-101 as the recognition network combined with a fully connected layer as a policy network.
ListenToLook uses MobileNet-V2 as the sampler network and ResNet-101 as the recognition network.
Finally, ListenToLook (IA$|$IA) uses two ResNet-18 for audio and visual modality respectively as the sampler. The same architecture is used for recognition network.

\begin{figure}[t!]
\centering
\includegraphics[width=1\linewidth]{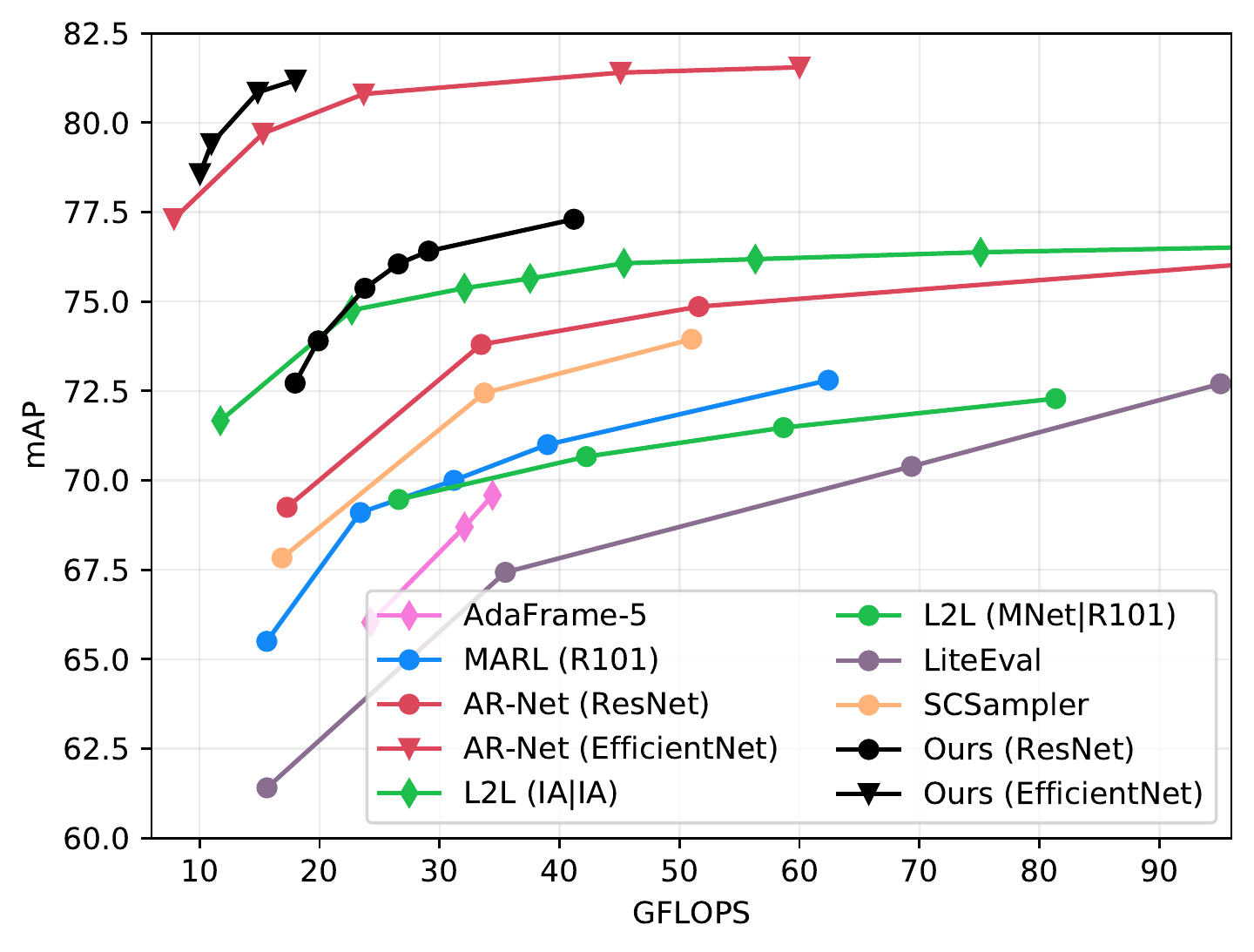}
\caption{\textbf{Accuracy vs. efficiency curves on ActivityNet.} \methodname~performs similar to or better than other methods at a much lower computational cost.
Note that L2L (IA$\vert$ IA) denotes the variant of the ListenToLook method that additionally uses audio for sampling and recognition.}
\label{fig:sota:AN}
\end{figure}

While the competing methods use two networks for sampling and recognizing, \methodname~uses a single network for efficient video recognition. The results for ActivityNet and Mini-Kinetics are shown in Table~\ref{table:sota}. Our method with a ResNet backbone outperforms all other approaches by obtaining an improved accuracy while using $1.3\times$-- $5\times$ less GFLOPs. The gain in accuracy is mainly attributed to our accumulated feature pooling module, while the gain in efficiency is attributed to the proposed sampling policy and gating modules for conditional early exiting (see Section~\ref{sec:exp:ablation} for detailed analyses). Compared to our model without early exiting, denoted as ``Ours (w/o exit)'', \methodname~achieves a comparable accuracy with a significant reduction in GFLOPs (see Appendix for wall-clock timing of \methodname). 
In addition, Figure~\ref{fig:sota:AN} presents accuracy vs. computations trade-off curves on ActivityNet for various methods. Note that except ListenToLook (IA|IA)~\cite{gao2020listen} that uses audio and visual features for sampling and recognition, other methods rely solely on visual features. As shown, \methodname~achieves the same top-performance as other methods at a much lower computational cost.

To verify that the performance of our model is not limited to one architecture, we conduct similar experiments with EfficientNet and X3D backbones. Using EfficientNet-b3, \methodname~obtains $3.9\%$ further absolute gain in mAP on ActivityNet and $2.5\%$ on Mini-Kinetics, while consuming $2.3\times$ and $2.5\times$ less compute. In particular, on ActivityNet and mini-kinetics, we outperform AR-Net~\cite{meng2020ar}, the leading method among competitors, with $1.3\times$ and $2.1\times$ less GFLOPs, respectively. When using the highly efficient X3D-S as our backbone, at a similar computational cost, FrameExit achieves 86.0\% which is 6.0\% higher than the best 2D model. This demonstrates the superiority of our method for efficient video understanding using both 2D and 3D backbones.

\medskip\partitle{Comparison to  state of the art: Holistic understanding.}
\label{sec:exp:hvu}
We evaluate our method for the task of holistic video understanding, where the goal is to recognize various semantic aspects including objects, scenes, and actions within a video.
To this end, we conduct experiments on the large-scale, multi-label HVU dataset~\cite{diba2020large}. we use average pooling in our accumulated feature pooling module, $\Psi$ in Eq.~\ref{eq:aggregation}, as it outperforms max pooling. We compare our method with the commonly-used uniform and random sampling baselines. These baselines sample $K$ frames (uniformly/randomly) and average frame-level predictions to generate a final video-level prediction. We additionally, compare our method with 3D-ResNet18~\cite{tran2017convnet} and mixed 2D/3D ResNet~\cite{diba2020large} models. As shown in Table~\ref{table:sota:hvu}, our method consistently outperforms other methods in terms of accuracy while saving significant computation. In particular, we report two variants of \methodname~with the ResNet backbone by changing accuracy-efficiency trade-off parameter $\beta$. \methodname~trained with $\beta=1e^{-3}$ only uses $8.6$ GFLOPs to obtain $45.7\%$ mAP, which is highly suitable for low-budget requirements. \methodname~trained with $\beta=1e^{-5}$ uses $18.7$ GFLOPs on average to obtain $49.2\%$ mAP, which is suitable for high-accuracy requirements. 
Interestingly, \methodname~outperforms 3D models~\cite{tran2017convnet, diba2020large} in this task. Given that HVU requires recognizing static 2D concepts such as scenes ($8\%$ of the class labels) and objects ($55\%$ of the class labels), 3D-CNN methods are not necessarily the optimal choice for this task. The most budget-friendly trained model is \methodname~with the EfficientNet backbone that uses only $5.7$ GFLOPs and achieves a mAP of $46.1$.

This experiment implies that our method can be used in a wider range of efficient video recognition tasks. Our results set a new state of the art on the HVU dataset for future research.

\begin{table}
\caption{\textbf{Comparison to state of the art on holistic video understand.} \methodname~ significantly improves mAP on HVU dataset while saving compute. 
$^*$indicates GFLOPs for one input clip.
}
\label{table:sota:hvu}
\centering
{
\begin{tabular}{l|cc}
\hline
 & mAP (\%) & GFLOPs\\
\hline
\rowcolor{mygray}\textit{2D/3D ResNet} & & \\
Uniform-10                              & $44.7$ & $41.2$\\
Random-10                               & $43.6$ & $41.2$\\
3D-ResNet18~\cite{tran2017convnet}      & $35.4$ & $38.6^*$\\
HATNet~\cite{diba2020large}             & $39.6$ & $41.8^*$\\

\textbf{\methodname} ($\beta=1e^{-3}$)  & $45.7$ & ${\bf 8.6}$\\
\textbf{\methodname} ($\beta=1e^{-5}$)  & ${\bf 49.2}$ & $18.7$\\

\hline
\rowcolor{mygray}\textit{EfficientNet} & & \\
\textbf{\methodname} ($\beta=1e^{-3}$) & $\bf{47.7}$ & $11.7$ \\
\textbf{\methodname} ($\beta=1e^{-2}$)  & $46.1$ & ${\bf 5.7}$ \\
\hline
\end{tabular}
}
\end{table}

\subsection{Ablation study}
\label{sec:exp:ablation}
In this section we inspect different aspects of \methodname. For all ablations, we follow the same training procedure explained in Section~\ref{sec:exp:implemetation_detail} and use the Activitynet-v1.3 dataset.

\medskip\partitle{Impact of policy function.}
To validate the impact of our proposed deterministic policy function for frame sampling, we make comparisons with two commonly used sampling approaches namely ``Sequential'' and ``Random'', both during training and evaluation. For ``Sequential'', we keep original frame order while in ``Random'', we randomly sample frames. Results for all combinations are shown in Table~\ref{table:ablation:ordering}. Using the original frame ordering during training and testing results in a lower mAP and a higher GFLOPs. We speculate that during training, earlier classifiers receive less informative gradients due to not observing a holistic view of the entire video (row 1 vs. row 2). Similarly, the model consumes higher GFLOPs because it needs to observe more frames to infer a confident prediction (row 1 vs. row 3). Random sampling both during training and testing improves the results because early stage classifiers have a higher chance of obtaining more useful information reflecting the entire video time-span. The best result, however, is obtained by our simple deterministic sampling policy. This is because jumping forward and backward enabled by our frame sampling policy effectively improves the chance of picking up informative frames for recognition.

\begin{table}[t!]
\caption{{Impact of sampling policy. we report the results over 5 runs}. 
}
\label{table:ablation:ordering}
\centering
\resizebox{1.0\columnwidth}{!}
{
\begin{tabular}{c|cc|cc}
\hline
 & Train-random & Test-random & mAP (\%) & GFLOPS\\
\hline
(1)                   & \xmark & \xmark & 73.7 & 27.6\\
(2)                   & \cmark & \xmark & 74.3 $\pm$ 0.9 & 26.7 $\pm$ 0.33 \\
(3)                   & \xmark & \cmark & 74.4 $\pm$ 0.02 & 27.0 $\pm$ 0.19\\
(4)                   & \cmark & \cmark & 74.4 $\pm$ 0.34 & 25.0 $\pm$ 0.37\\
\methodname          & - & - &  76.1 & 26.1 \\
\hline
\end{tabular}
}
\end{table}

\medskip\partitle{Impact of accumulated feature pooling.}
\begin{table}[b]
\caption{{Impact of accumulated feature pooling.}}
\label{table:ablation:pooling}
\centering
{
\begin{tabular}{c|cc}
\hline
Feature pooling & mAP (\%) & GFLOPs\\
\hline
\cmark & 76.1 & 26.1 \\
\xmark & 67.8 & 27.6 \\
\hline
\end{tabular}
}
\end{table}
We evaluate the impact of accumulated feature pooling module by comparing the performance of \methodname~with and without feature pooling as reported in Table~\ref{table:ablation:pooling}.
For the \methodname~ without feature pooling, we train each classifier and its associated gating module only on the currently sampled frame. We then average frame-level predictions made by all classifiers up until the exiting classifier to obtain the video-level prediction. Therefore, the major difference in the design of these two settings relates to the use of the pooling operation in feature space or in the prediction space. The results in Table~\ref{table:ablation:pooling} demonstrates that pooling over features is much more effective than over output predictions. 

\medskip\partitle{Number of input frames.}
We train \methodname~ with various number of input frames $T=\{4,8,10,16\}$. During inference, the gates decide when to stop processing. As shown in Table~\ref{table:ablation:K}, the performance of the model increases as the number of input frames increases, but up to a certain limit. This is a reasonable observation, as certain actions/videos may require more frames to be recognized. However, the reason why increasing the number of frames after a limit does not further improve the performance could be mainly attributed to the limited capacity of 2D convolutional networks in leveraging temporal information. As a result, to keep the balance between accuracy and efficiency, we set $T=10$ in our experiments. 

\begin{table}[t]
\caption{{Impact of number of frames on \methodname}. 
}
\label{table:ablation:K}
\centering
{
\begin{tabular}{c|cccc}
\hline
 & $T=4$ & $T=8$ & $T=10$ & $T=16$\\
\hline
mAP (\%)    & $66.2$ & $74.3$ & $76.1$ & $76.1$ \\ 
GFLOPs      & $13.0$ & $22.4$ & $26.1$ & $35.1$ \\
\hline
\end{tabular}
}
\end{table}


\medskip\partitle{Adaptive vs fixed exiting.}
To show the merits of adaptive early exiting versus a fixed budget exiting, we conduct an ablation in two settings. In the first setting, we use our conditional early exiting model and in the second setting, we assume a fixed number of frames is processed for each test video. Table~\ref{table:ablation:dynamic} shows that our conditional early exiting method consistently outperforms fixed exiting. 

\begin{table}[t]
\caption{{Impact of adaptive exiting}. $^*$for each column, results are reported over an average no. of processed frames.
}
\label{table:ablation:dynamic}
\centering
{
\begin{tabular}{c|cccc}
\hline
No. processed frames & $3$ & $4$ & $6$ & $7$\\
\hline
Fixed budget     & $60.8$ & $67.5$ & $74.0$ & $75.2$ \\ 
\methodname*    & $67.5$ & $70.1$ & $75.8$ & $76.4$ \\
\hline
\end{tabular}
}
\end{table}


\section{Conclusions}
In this paper, we presented \methodname, a conditional early exiting method for efficient video recognition. Our proposed method uses gating modules, trained to allow the network to automatically determine the earliest exiting point based on the inferred complexity of the input video. To enable gates to make reliable decisions we use an effective video representation obtained using accumulated feature pooling. 
We showed that our early exiting mechanism combined with a simple, deterministic sampling strategy obviates the need for complex sampling policy techniques.
Our proposed method is model-agnostic and can be used with various network architectures. Comprehensive experiments show the power of our method in balancing accuracy versus efficiency in action recognition as well as holistic video understanding tasks.

\paragraph{Acknowledgements} We thank Fatih Porikli, Michael Hofmann, Haitam Ben Yahia, Arash Behboodi and Ilia Karmanov for their feedback and discussions.



{\small
\bibliographystyle{ieee_fullname}
\bibliography{main}

\begin{thebibliography}{10}\itemsep=-1pt

\bibitem{abati2020conditional}
Davide Abati, Jakub Tomczak, Tijmen Blankevoort, Simone Calderara, Rita
  Cucchiara, and Babak~Ehteshami Bejnordi.
\newblock Conditional channel gated networks for task-aware continual learning.
\newblock In {\em Proceedings of the IEEE/CVF Conference on Computer Vision and
  Pattern Recognition}, pages 3931--3940, 2020.

\bibitem{bejnordi2019batch}
Babak~Ehteshami Bejnordi, Tijmen Blankevoort, and Max Welling.
\newblock Batch-shaping for learning conditional channel gated networks.
\newblock In {\em International Conference on Learning Representations}, 2019.

\bibitem{caba2015activitynet}
Fabian Caba~Heilbron, Victor Escorcia, Bernard Ghanem, and Juan Carlos~Niebles.
\newblock Activitynet: A large-scale video benchmark for human activity
  understanding.
\newblock In {\em Proceedings of the ieee conference on computer vision and
  pattern recognition}, pages 961--970, 2015.

\bibitem{campos2017skip}
V{\'\i}ctor Campos, Brendan Jou, Xavier Gir{\'o}-i Nieto, Jordi Torres, and
  Shih-Fu Chang.
\newblock Skip rnn: Learning to skip state updates in recurrent neural
  networks.
\newblock {\em arXiv preprint arXiv:1708.06834}, 2017.

\bibitem{carreira2017quo}
Joao Carreira and Andrew Zisserman.
\newblock Quo vadis, action recognition? a new model and the kinetics dataset.
\newblock In {\em CVPR}, 2017.

\bibitem{gaternet}
Zhourong Chen, Yang Li, Samy Bengio, and Si Si.
\newblock Gaternet: Dynamic filter selection in convolutional neural network
  via a dedicated global gating network.
\newblock {\em arXiv preprint arXiv:1811.11205}, 2018.

\bibitem{diba2020large}
Ali Diba, Mohsen Fayyaz, Vivek Sharma, Manohar Paluri, J{\"u}rgen Gall, Rainer
  Stiefelhagen, and Luc Van~Gool.
\newblock Large scale holistic video understanding.
\newblock In {\em European Conference on Computer Vision}, pages 593--610.
  Springer, 2020.

\bibitem{fan2018watching}
Hehe Fan, Zhongwen Xu, Linchao Zhu, Chenggang Yan, Jianjun Ge, and Yi Yang.
\newblock Watching a small portion could be as good as watching all: Towards
  efficient video classification.
\newblock In {\em IJCAI International Joint Conference on Artificial
  Intelligence}, 2018.

\bibitem{fan2020rubiksnet}
Linxi Fan, Shyamal Buch, Guanzhi Wang, Ryan Cao, Yuke Zhu, Juan~Carlos Niebles,
  and Li Fei-Fei.
\newblock Rubiksnet: Learnable 3d-shift for efficient video action recognition.
\newblock In {\em European Conference on Computer Vision}, pages 505--521.
  Springer, 2020.

\bibitem{fan2019more}
Quanfu Fan, Chun-Fu Chen, Hilde Kuehne, Marco Pistoia, and David Cox.
\newblock More is less: Learning efficient video representations by big-little
  network and depthwise temporal aggregation.
\newblock {\em arXiv preprint arXiv:1912.00869}, 2019.

\bibitem{feichtenhofer2020x3d}
Christoph Feichtenhofer.
\newblock X3d: Expanding architectures for efficient video recognition.
\newblock In {\em Proceedings of the IEEE/CVF Conference on Computer Vision and
  Pattern Recognition}, pages 203--213, 2020.

\bibitem{feichtenhofer2019slowfast}
Christoph Feichtenhofer, Haoqi Fan, Jitendra Malik, and Kaiming He.
\newblock Slowfast networks for video recognition.
\newblock In {\em Proceedings of the IEEE international conference on computer
  vision}, pages 6202--6211, 2019.

\bibitem{fernando2016rank}
Basura Fernando, Efstratios Gavves, Jos{\'e} Oramas, Amir Ghodrati, and Tinne
  Tuytelaars.
\newblock Rank pooling for action recognition.
\newblock {\em IEEE transactions on pattern analysis and machine intelligence},
  39(4):773--787, 2016.

\bibitem{figurnov2017spatially}
Michael Figurnov, Maxwell~D Collins, Yukun Zhu, Li Zhang, Jonathan Huang,
  Dmitry Vetrov, and Ruslan Salakhutdinov.
\newblock Spatially adaptive computation time for residual networks.
\newblock In {\em CVPR}, 2017.

\bibitem{gao2020listen}
Ruohan Gao, Tae-Hyun Oh, Kristen Grauman, and Lorenzo Torresani.
\newblock Listen to look: Action recognition by previewing audio.
\newblock In {\em Proceedings of the IEEE/CVF Conference on Computer Vision and
  Pattern Recognition}, pages 10457--10467, 2020.

\bibitem{dynamicchannelpruning}
Xitong Gao, Yiren Zhao, Lukasz Dudziak, Robert Mullins, and Cheng{-}Zhong Xu.
\newblock Dynamic channel pruning: Feature boosting and suppression.
\newblock {\em arXiv preprint arxiv:1810.05331}, 2018.

\bibitem{ghodrati2018video}
Amir Ghodrati, Efstratios Gavves, and Cees~GM Snoek.
\newblock Video time: Properties, encoders and evaluation.
\newblock {\em arXiv preprint arXiv:1807.06980}, 2018.

\bibitem{hara2017learning}
Kensho Hara, Hirokatsu Kataoka, and Yutaka Satoh.
\newblock Learning spatio-temporal features with 3d residual networks for
  action recognition.
\newblock In {\em Proceedings of the IEEE International Conference on Computer
  Vision Workshops}, pages 3154--3160, 2017.

\bibitem{he2016deep}
Kaiming He, Xiangyu Zhang, Shaoqing Ren, and Jian Sun.
\newblock Deep residual learning for image recognition.
\newblock In {\em Proceedings of the IEEE conference on computer vision and
  pattern recognition}, 2016.

\bibitem{hochreiter1997long}
Sepp Hochreiter and J{\"u}rgen Schmidhuber.
\newblock Long short-term memory.
\newblock {\em Neural computation}, 9(8):1735--1780, 1997.

\bibitem{yt_video_qualitative4}
\href{https://www.youtube.com/watch?v=0lkvs1QUcjI}{ORBEA ALMA M30 CARBON ·
  2014 / BICIMAG} by
  \href{https://www.youtube.com/channel/UCNHEGHe18bf7cTbjHzbgO8g}{bicimag} is
  licensed~under
  \href{https://creativecommons.org/licenses/by/3.0/legalcode}{CC BY}.

\bibitem{yt_video_qualitative1}
\href{https://www.youtube.com/watch?v=0wHOYxjRmlw}{Sophie and Greg playing acro
  yoga} by
  \href{https://www.youtube.com/channel/UCRoRt4xCgMd2O8ykNoN5boA}{Gregology} is
  licensed~under
  \href{https://creativecommons.org/licenses/by/3.0/legalcode}{CC BY}.

\bibitem{yt_video_qualitative3}
\href{https://www.youtube.com/watch?v=5-3GN1nFB60}{Kilby Plays the Harmonica -
  Sort of} by
  \href{https://www.youtube.com/channel/UC91ZM2e4B6OroBcjMTyhR_A}{aplusjimages}
  is licensed~under
  \href{https://creativecommons.org/licenses/by/3.0/legalcode}{CC BY}.

\bibitem{yt_video_teaser1}
\href{https://www.youtube.com/watch?v=62IHefGRNPc}{SPORK! Exclusive: Pathways
  Waveland Bowling} by
  \href{https://www.youtube.com/channel/UCHE4991XFTTonXsv4vHfEpQ}{SPORK! NFP}
  is licensed~under
  \href{https://creativecommons.org/licenses/by/3.0/legalcode}{CC BY}.

\bibitem{yt_video_teaser3}
\href{https://www.youtube.com/watch?v=buADPfbPNkY}{Canada's National Ballet
  School: Training in the Professional Ballet Program} by
  \href{https://www.youtube.com/channel/UCoG4XLDqoaPGWpNXFBhrYkg}{Canada's
  National Ballet School} is licensed~under
  \href{https://creativecommons.org/licenses/by/3.0/legalcode}{CC BY}.

\bibitem{yt_video_qualitative5}
\href{https://www.youtube.com/watch?v=FLUTSVeTZGg}{Batlle of The Year: Dau
  Truong Breakdance TV Spot} by
  \href{https://www.youtube.com/channel/UCq_28uJmbq4tlji7YAMZGtg}{Galaxy
  Studio} is licensed~under
  \href{https://creativecommons.org/licenses/by/3.0/legalcode}{CC BY}.

\bibitem{yt_video_qualitative2}
\href{https://www.youtube.com/watch?v=TcXkhoyfULU}{Show de tango COOPRUDEA} by
  \href{https://www.youtube.com/channel/UC8n-UdYvytHoxh97PmXP2jQ}{Cooperativa
  de Profesores Cooprudea} is licensed~under
  \href{https://creativecommons.org/licenses/by/3.0/legalcode}{CC BY}.

\bibitem{multiscaledense}
Gao Huang, Danlu Chen, Tianhong Li, Felix Wu, Laurens van~der Maaten, and
  Kilian Weinberger.
\newblock Multi-scale dense networks for resource efficient image
  classification.
\newblock In {\em International Conference on Learning Representations}, 2018.

\bibitem{hussein2020timegate}
Noureldien Hussein, Mihir Jain, and Babak~Ehteshami Bejnordi.
\newblock Timegate: Conditional gating of segments in long-range activities.
\newblock {\em arXiv preprint arXiv:2004.01808}, 2020.

\bibitem{kay2017kinetics}
Will Kay, Joao Carreira, Karen Simonyan, Brian Zhang, Chloe Hillier, Sudheendra
  Vijayanarasimhan, Fabio Viola, Tim Green, Trevor Back, Paul Natsev, et~al.
\newblock The kinetics human action video dataset.
\newblock {\em arXiv preprint arXiv:1705.06950}, 2017.

\bibitem{adam}
Diederik~P. Kingma and Jimmy Ba.
\newblock Adam: A method for stochastic optimization.
\newblock In {\em ICLR}, 2014.

\bibitem{kopuklu2019resource}
Okan K{\"o}p{\"u}kl{\"u}, Neslihan Kose, Ahmet Gunduz, and Gerhard Rigoll.
\newblock Resource efficient 3d convolutional neural networks.
\newblock In {\em 2019 IEEE/CVF International Conference on Computer Vision
  Workshop (ICCVW)}, pages 1910--1919. IEEE, 2019.

\bibitem{korbar2019scsampler}
Bruno Korbar, Du Tran, and Lorenzo Torresani.
\newblock Scsampler: Sampling salient clips from video for efficient action
  recognition.
\newblock In {\em Proceedings of the IEEE International Conference on Computer
  Vision}, pages 6232--6242, 2019.

\bibitem{lee2015deeply}
Chen-Yu Lee, Saining Xie, Patrick Gallagher, Zhengyou Zhang, and Zhuowen Tu.
\newblock Deeply-supervised nets.
\newblock In {\em Artificial intelligence and statistics}, pages 562--570,
  2015.

\bibitem{lin2019tsm}
Ji Lin, Chuang Gan, and Song Han.
\newblock Tsm: Temporal shift module for efficient video understanding.
\newblock In {\em Proceedings of the IEEE International Conference on Computer
  Vision}, pages 7083--7093, 2019.

\bibitem{maddison2016concrete}
Chris~J Maddison, Andriy Mnih, and Yee~Whye Teh.
\newblock The concrete distribution: A continuous relaxation of discrete random
  variables.
\newblock In {\em ICLR}, 2017.

\bibitem{meng2020ar}
Yue Meng, Chung-Ching Lin, Rameswar Panda, Prasanna Sattigeri, Leonid
  Karlinsky, Aude Oliva, Kate Saenko, and Rogerio Feris.
\newblock Ar-net: Adaptive frame resolution for efficient action recognition.
\newblock {\em arXiv preprint arXiv:2007.15796}, 2020.

\bibitem{meng2021adafuse}
Yue Meng, Rameswar Panda, Chung-Ching Lin, Prasanna Sattigeri, Leonid
  Karlinsky, Kate Saenko, Aude Oliva, and Rogerio Feris.
\newblock Adafuse: Adaptive temporal fusion network for efficient action
  recognition.
\newblock In {\em International Conference on Learning Representations}, 2021.

\bibitem{piergiovanni2019tiny}
AJ Piergiovanni, Anelia Angelova, and Michael~S Ryoo.
\newblock Tiny video networks.
\newblock {\em arXiv preprint arXiv:1910.06961}, 2019.

\bibitem{qiu2017learning}
Zhaofan Qiu, Ting Yao, and Tao Mei.
\newblock Learning spatio-temporal representation with pseudo-3d residual
  networks.
\newblock In {\em proceedings of the IEEE International Conference on Computer
  Vision}, pages 5533--5541, 2017.

\bibitem{sutton2018reinforcement}
Richard~S Sutton and Andrew~G Barto.
\newblock {\em Reinforcement learning: An introduction}.
\newblock MIT press, 2018.

\bibitem{tan2019efficientnet}
Mingxing Tan and Quoc~V Le.
\newblock Efficientnet: Rethinking model scaling for convolutional neural
  networks.
\newblock {\em arXiv preprint arXiv:1905.11946}, 2019.

\bibitem{branchynet}
S. {Teerapittayanon}, B. {McDanel}, and H.~T. {Kung}.
\newblock Branchynet: Fast inference via early exiting from deep neural
  networks.
\newblock In {\em 2016 23rd International Conference on Pattern Recognition
  (ICPR)}, pages 2464--2469, Dec 2016.

\bibitem{tran2015learning}
Du Tran, Lubomir Bourdev, Rob Fergus, Lorenzo Torresani, and Manohar Paluri.
\newblock Learning spatiotemporal features with 3d convolutional networks.
\newblock In {\em Proceedings of the IEEE international conference on computer
  vision}, pages 4489--4497, 2015.

\bibitem{tran2017convnet}
Du Tran, Jamie Ray, Zheng Shou, Shih-Fu Chang, and Manohar Paluri.
\newblock Convnet architecture search for spatiotemporal feature learning.
\newblock {\em arXiv preprint arXiv:1708.05038}, 2017.

\bibitem{tran2019video}
Du Tran, Heng Wang, Lorenzo Torresani, and Matt Feiszli.
\newblock Video classification with channel-separated convolutional networks.
\newblock In {\em Proceedings of the IEEE International Conference on Computer
  Vision}, pages 5552--5561, 2019.

\bibitem{tran2018closer}
Du Tran, Heng Wang, Lorenzo Torresani, Jamie Ray, Yann LeCun, and Manohar
  Paluri.
\newblock A closer look at spatiotemporal convolutions for action recognition.
\newblock In {\em Proceedings of the IEEE conference on Computer Vision and
  Pattern Recognition}, pages 6450--6459, 2018.

\bibitem{vaswani2017attention}
Ashish Vaswani, Noam Shazeer, Niki Parmar, Jakob Uszkoreit, Llion Jones,
  Aidan~N Gomez, {\L}ukasz Kaiser, and Illia Polosukhin.
\newblock Attention is all you need.
\newblock In {\em Advances in neural information processing systems}, pages
  5998--6008, 2017.

\bibitem{convnetaig}
Andreas Veit and Serge Belongie.
\newblock Convolutional networks with adaptive inference graphs.
\newblock In {\em European Conference on Computer Vision (ECCV)}, 2018.

\bibitem{verelst2020dynamic}
Thomas Verelst and Tinne Tuytelaars.
\newblock Dynamic convolutions: Exploiting spatial sparsity for faster
  inference.
\newblock In {\em Proceedings of the IEEE/CVF Conference on Computer Vision and
  Pattern Recognition}, pages 2320--2329, 2020.

\bibitem{wang2020learning}
Longguang Wang, Xiaoyu Dong, Yingqian Wang, Xinyi Ying, Zaiping Lin, Wei An,
  and Yulan Guo.
\newblock Learning sparse masks for efficient image super-resolution.
\newblock {\em arXiv preprint arXiv:2006.09603}, 2020.

\bibitem{skipnet}
Xin Wang, Fisher Yu, Zi{-}Yi Dou, Trevor Darrell, and Joseph~E. Gonzalez.
\newblock Skipnet: Learning dynamic routing in convolutional networks.
\newblock In {\em Computer Vision - {ECCV} 2018 - 15th European Conference,
  Munich, Germany, September 8-14, 2018, Proceedings, Part {XIII}}, pages
  420--436, 2018.

\bibitem{wu2019multi}
Wenhao Wu, Dongliang He, Xiao Tan, Shifeng Chen, and Shilei Wen.
\newblock Multi-agent reinforcement learning based frame sampling for effective
  untrimmed video recognition.
\newblock In {\em Proceedings of the IEEE/CVF International Conference on
  Computer Vision}, pages 6222--6231, 2019.

\bibitem{wu2018blockdrop}
Zuxuan Wu, Tushar Nagarajan, Abhishek Kumar, Steven Rennie, Larry~S Davis,
  Kristen Grauman, and Rogerio Feris.
\newblock Blockdrop: Dynamic inference paths in residual networks.
\newblock In {\em Proceedings of the IEEE Conference on Computer Vision and
  Pattern Recognition}, pages 8817--8826, 2018.

\bibitem{wu2019liteeval}
Zuxuan Wu, Caiming Xiong, Yu-Gang Jiang, and Larry~S Davis.
\newblock Liteeval: A coarse-to-fine framework for resource efficient video
  recognition.
\newblock In {\em Advances in Neural Information Processing Systems}, pages
  7780--7789, 2019.

\bibitem{wu2019adaframe}
Zuxuan Wu, Caiming Xiong, Chih-Yao Ma, Richard Socher, and Larry~S Davis.
\newblock Adaframe: Adaptive frame selection for fast video recognition.
\newblock In {\em Proceedings of the IEEE Conference on Computer Vision and
  Pattern Recognition}, pages 1278--1287, 2019.

\bibitem{xie2017rethinking}
Saining Xie, Chen Sun, Jonathan Huang, Zhuowen Tu, and Kevin Murphy.
\newblock Rethinking spatiotemporal feature learning for video understanding.
\newblock {\em arXiv preprint arXiv:1712.04851}, 1(2):5, 2017.

\bibitem{yang2020resolution}
Le Yang, Yizeng Han, Xi Chen, Shiji Song, Jifeng Dai, and Gao Huang.
\newblock Resolution adaptive networks for efficient inference.
\newblock In {\em Proceedings of the IEEE/CVF Conference on Computer Vision and
  Pattern Recognition}, pages 2369--2378, 2020.

\bibitem{yeung2016end}
Serena Yeung, Olga Russakovsky, Greg Mori, and Li Fei-Fei.
\newblock End-to-end learning of action detection from frame glimpses in
  videos.
\newblock In {\em Proceedings of the IEEE Conference on Computer Vision and
  Pattern Recognition}, pages 2678--2687, 2016.

\bibitem{yue2015beyond}
Joe Yue-Hei~Ng, Matthew Hausknecht, Sudheendra Vijayanarasimhan, Oriol Vinyals,
  Rajat Monga, and George Toderici.
\newblock Beyond short snippets: Deep networks for video classification.
\newblock In {\em Proceedings of the IEEE conference on computer vision and
  pattern recognition}, pages 4694--4702, 2015.

\bibitem{zheng2020dynamic}
Yin-Dong Zheng, Zhaoyang Liu, Tong Lu, and Limin Wang.
\newblock Dynamic sampling networks for efficient action recognition in videos.
\newblock {\em IEEE Transactions on Image Processing}, 29:7970--7983, 2020.

\bibitem{zhou2018temporal}
Bolei Zhou, Alex Andonian, Aude Oliva, and Antonio Torralba.
\newblock Temporal relational reasoning in videos.
\newblock In {\em Proceedings of the European Conference on Computer Vision
  (ECCV)}, pages 803--818, 2018.

\bibitem{zolfaghari2018eco}
Mohammadreza Zolfaghari, Kamaljeet Singh, and Thomas Brox.
\newblock Eco: Efficient convolutional network for online video understanding.
\newblock In {\em Proceedings of the European conference on computer vision
  (ECCV)}, pages 695--712, 2018.

\end{thebibliography}
}

\vspace{5mm}
\section*{Appendix}
\vspace{5mm}
\subsection*{A. Algorithmic description of the policy}
In this section, we present an algorithmic description of our policy function. Our policy function follows a coarse-to-fine principle for sampling frames. It starts sampling from a coarse temporal scale and gradually samples more frames to add finer details to the temporal structure. Specifically, we sample the first frames from the middle, beginning, and end of the video, respectively, and then repeatedly sample frames from the halves as shown in Algorithm~\ref{alg:policy}.

\SetKwInput{KwResult}{Output}
\SetKwInput{KwInput}{Input}

\begin{algorithm}
\SetAlgoLined
\KwInput{N (number of frames in the video)}
\KwResult{\textbf{S} (a sequence of sampled frame indices)}
 \textbf{S} = [floor((N+1)/2), 1, N] \;
 q = 2 \;
 \While{len(\textbf{S}) $<$ N}{
      interval = floor(linspace(1, N, q+1))\;
      \For{i=1 : len(interval)-1}{
        a = interval[i]\;
        b = interval[i+1]\;
        ind = floor((a+b)/2)\;
        \If{ind is not in \textbf{S}}{
            \textbf{S}.append(ind)\;
        }
      }
      q = q * 2\;
 }
\caption{Policy function}\label{alg:policy}
\end{algorithm}

\subsection*{B. Wall-clock time inference}
The wall-clock speedup of FrameExit is highly correlated with FLOPS. This is because the major computational cost comes from the number of frames being processed by the backbone. For example, using an Nvidia GeForce GTX 1080Ti GPU, the inference time of FrameExit with a ResNet50 backbone at computational costs of 41.2, 26.1, and 12.3 GFLOPs are 15.8, 10.9, and 4.8 ms, respectively.

\subsection*{C. Qualitative results}
\begin{figure}[h]
\centering
\includegraphics[width=1\linewidth]{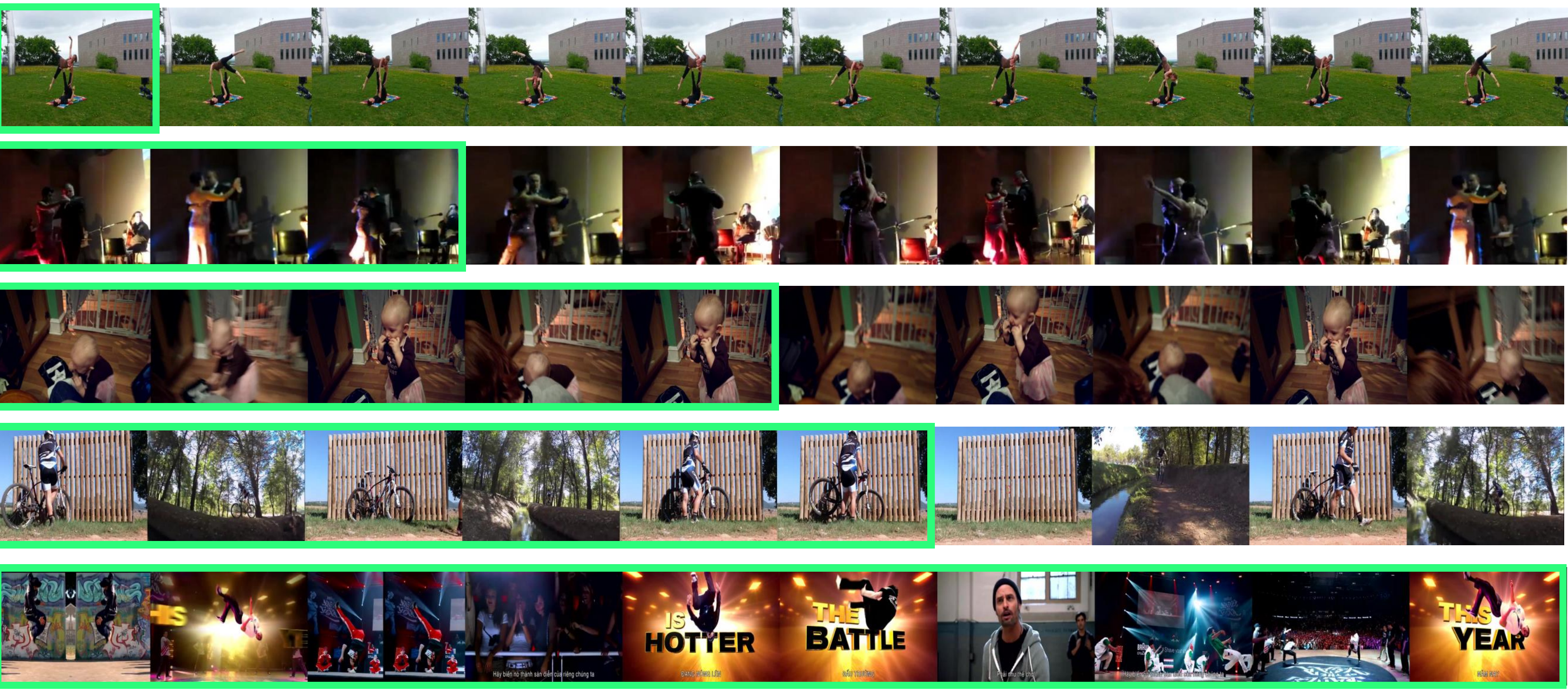}
\caption{\textbf{Qualitative results.} Each row contains $10$ frames of a video, sampled according to the policy function. Our method observes only the green box to recognize an action. As the content becomes more complex, \methodname~needs more frames to make a confident prediction. Zoom in for higher quality. Videos are adopted from ~\cite{yt_video_qualitative4, yt_video_qualitative1, yt_video_qualitative3, yt_video_qualitative5, yt_video_qualitative2}.}
\label{fig:qualitative}
\end{figure}

Figure~\ref{fig:qualitative} presents several video clips and the number of frames required to process by \methodname~before exiting. As shown, a few frames are sufficient for \methodname~to classify a video if the observed frames are discriminative enough. As the complexity of video scene increases (such as obscured objects and cluttered background) \methodname~needs more temporal information to reach a confident prediction.

\vspace{-3mm}

\end{document}